# A Multi-criteria neutrosophic group decision making metod based TOPSIS for supplier selection

*Rıdvan şahin\* and Muhammed Yiğider\*\**

*\*Department of Mathematics, Faculty of Science, Ataturk University, Erzurum*
*\*\*Department of Mathematics, Faculty of Science, Erzurum Technical University, Erzurum*

mat.ridone@gmail.com
myigider@gmail.com

**Abstract**

The process of multiple criteria decision making (MCDM) is of determining the best choice among all of the probable alternatives. The problem of supplier selection on which decision maker has usually vague and imprecise knowledge is a typical example of multi criteria group decision-making problem. The conventional crisp techniques has not much effective for solving MCDM problems because of imprecise or fuzziness nature of the linguistic assessments. To find the exact values for MCDM problems is both difficult and impossible in more cases in real world. So, it is more reasonable to consider the values of alternatives according to the criteria as single valued neutrosophic sets (SVNS). This paper deal with the technique for order preference by similarity to ideal solution (TOPSIS) approach and extend the TOPSIS method to MCDM problem with single valued neutrosophic information. The value of each alternative and the weight of each criterion are characterized by single valued neutrosophic numbers. Here, the importance of criteria and alternatives is identified by aggregating individual opinions of decision makers (DMs) via single valued neutrosophic weighted averaging (IFWA) operator. The proposed method is, easy use, precise and practical for solving MCDM problem with single valued neutrosophic data. Finally, to show the applicability of the developed method, a numerical experiment for supplier choice is given as an application of single valued neutrosophic TOPSIS method at end of this paper.

## 1. Introduction

The concept of neutrosophic set (NS) developed by Smarandache ([1,2]) is a more general platform which extends the concepts of the classic set and fuzzy set ([3]), intuitionistic fuzzy set ([4]) and interval valued intuitionistic fuzzy sets ([5]). In contrast to intuitionistic fuzzy sets and also interval valued intuitionistic fuzzy sets, the indeterminacy is characterized explicitly in a neutrosophic set. A neutrosophic set has three basic components such that truth membership $T$, indeterminacy membership $I$ and falsity membership $F$, which are defined independently of one another. But, a neutrosophic set will be more difficult to apply in real scientific and engineering fields. Therefore, Wang et al. ([6,7]) proposed the concepts of single valued neutrosophic set (SVNS) and interval neutrosophic set (INS) which are an instance of a neutrosophic set, and provided the set-theoretic operators and various properties of SVNSs and INSs. SVNSs present uncertainty, imprecise, inconsistent and incomplete information existing in real world. Also, it would be more suitable to handle indeterminate information and inconsistent information.

We usually need the decision making methods because of the uncertainty and complex under the physical nature of the problems. By the multi-criteria decision making methods, we can determine the best alternative from multiple alternatives with respect to some criteria. Recently, supplier selection has become increasingly important in both academia and industry (see [8-12]). So there are many MCDM techniques developed for the supplier selection problem. Some of these techniques are categorical method, weighted point method ([13]), matrix approach ([14]), vendor performance matrix approach ([15]) vendor profile analysis (VPA) ([16]) analytic hierarchy process (AHP) ([17-19]), analytic network process (ANP) ([20]), mathematical programming ([21,22]) and multiple objective programming (MOP) ([23-25]). However, most of these methods are developed with respect to crisp data and so they have not several influence factors such as imprecision preferences, additional qualitative criteria and incomplete information. Therefore, fuzzy set theory (FST) is more appropriate to overcome problems in decision making process.

Li et al. ([26]) and Holt ([27]) proposed the application of supplied selection under fuzzy data. Chen et al. ([28]) extended the concept of classic

Keywords: Neutrosophic sets, single valued neutrosophic sets, TOPSIS method, multi-criteria decision making, distance measure

TOPSIS method to solve supplier selection problems in fuzzy set theory.

TOPSIS (Technique for order preference by similarity to an ideal solution) method which is one of the most used classical MCDM methods has developed by Hwang and Yoon ([29]). Then the proposed set theories have provided the different multi-criteria decision making methods. Some authors ([30-41]) studied on multi-criteria decision-making methods based fuzzy data. Boran et al. ([41]) proposed the TOPSIS method to select appropriate supplier under intuitionistic fuzzy environment. Then the TOPSIS method for MCDM problem has extended in interval valued intuitionistic fuzzy sets by Ye ([42]).

As mentioned above, the single valued neutrosophic information is a generalization of intuitionistic fuzzy information, while intuitionistic fuzzy information is generalizes the fuzzy information. On one hand, a single valued neutrosophic set is an instance of neutrosophic set, which give us an additional possibility to represent uncertainty, imprecise, incomplete, and inconsistent information existing in real world. It can describe and handle indeterminate information and inconsistent information. However, the connector in the fuzzy set is defined with respect to T, i.e. membership only, hence the information of indeterminacy and non-membership is lost. The connectors in the intuitionistic fuzzy set are defined with respect to T and F, i.e. membership and non-membership only, hence the indeterminacy is what is left from 1, while in the neutrosophic set, they can be defined by any of them (no restriction) ([1]). For example, when we ask the opinion of an expert about certain statement, one may say that the possibility in which the statement is true is 0.6, the statement is false is 0.5 and the statement is not sure is 0.2. For neutrosophic notation, it can be expressed as $x(0.6,0.2,0.5)$. For further example, suppose there are 10 voters during a voting process. Five vote "aye", two vote "blackball" and three vote are undecided. For neutrosophic notation, it can be characterized as $x(0.5,0.3,0.2)$. However, the expression are beyond the scope of the intuitionistic fuzzy set. Therefore, the concept of single valued neutrosophic set is more general structure and very suitable to overcome the mentioned issues. Then we say that the TOPSIS method under single valued neutrosophic environment is suitable for decision making. Moreover, the single valued neutrosophic TOPSIS not only use for single valued neutrosophic information, but also extends the intuitionistic fuzzy TOPSIS and the fuzzy TOPSIS.

But, until now there have been no many studies on multi criteria decision making methods which are criterion values for alternatives are single valued neutrosophic sets. Ye ([43]) presented the correlation coefficient of SVNSs and the cross-entropy measure of SVNSs and applied them to single valued neutrosophic decision-making problems. Also Ye ([44]) defined single valued neutrosophic cross entropy which is proposed as an extension of the cross entropy of fuzzy sets, Recently, Zhang et al. ([46]) established two interval neutrosophic aggregation operators such as interval neutrosophic weighted arithmetic operator and interval neutrosophic weighted geometric operator and presented a method for multi criteria decision making problems based on the aggregation operators.

The main purposes of this paper were (1) to define one equation to calculate performance weights of decision makers expressed by single valued neutrosophic numbers (2) to establish a multi criteria decision making method based on TOPSIS method under single valued neutrosophic values for supplier selection, (3) to show the application and effectiveness of the proposed method with an example, and (4) to present performances of alternatives according to each criterion via graphics visualizing the relationships among alternatives, SVN positive ideal solution and SVN negative ideal solution.

We organize the rest of the paper as follows: in the following section, we give preliminary definitions of neutrosophic sets and single valued neutrosophic sets and propose a score function for ranking SVN numbers. In Section 3, we present a technical to extend TOPSIS method in single valued neutrosophic environment. In Section 4, we illustrate our developed method by an example. This paper is terminated in Section 5.

## 2. Preliminaries

### 2.1 Neutrosophic set

In the following, we give a brief review of some preliminaries.

**Definition 1.** [1] Let $X$ be a space of points (objects) and $x \in X$. A neutrosophic set $A$ in $X$ is defined by a truth-membership function $T_A(x)$, an indeterminacy-membership function $I_A(x)$ and a falsity-membership function $F_A(x)$. $T_A(x), I_A(x)$ and $F_A(x)$ are real standard or real nonstandard subsets of $]0^-, 1^+[$. That is $T_A(x): X \to ]0^-, 1^+[$, $T_A(x): X \to ]0^-, 1^+[$ and $T_A(x): X \to ]0^-, 1^+[$. There is not restriction on the sum of $T_A(x), I_A(x)$ and $F_A(x)$, so $0^- \leq \sup T_A(x) \leq \sup I_A(x) \leq \sup F_A(x) \leq 3^+$.

In the following, we adopt the notations $u_A(x), r_A(x)$ and $v_A(x)$ instead of $T_A(x), I_A(x)$ and $F_A(x)$, respectively. Also we write SVN numbers instead of single valued neutrosophic numbers.

*2.2 Single valued neutrosophic sets*

A single valued neutrosophic set (SVNS) has been defined in ([6]) as follows:

**Definition 2.** Let $X$ be a universe of discourse. A single valued neutrosophic set $A$ over $X$ is an object having the form

$$A = \{\langle x, u_A(x), r_A(x), v_A(x)\rangle : x \in X\} \quad (1)$$

where $u_A(x): X \to [0,1]$, $r_A(x): X \to [0,1]$ and $v_A(x): X \to [0,1]$ with $0 \leq u_A(x) + r_A(x) + v_A(x) \leq 3$ for all $x \in X$. The intervals $u_A(x), r_A(x)$ and $v_A(x)$ denote the truth- membership degree, the indeterminacy-membership degree and the falsity membership degree of $x$ to $A$, respectively.

For convenience, a SVN number is denoted by $A = (a, b, c)$, where $a, b, c \in [0,1]$ and $a + b + c \leq 3$.

**Definition 3.** [46] Let $A_1 = (a_1, b_1, c_1)$ and $A_2 = (a_2, b_2, c_2)$ be two SVN numbers, then summation between $A_1$ and $A_2$ is defined as follows:

$$A_1 \oplus A_2 = (a_1 + a_2 - a_1 a_2, b_1 b_2, c_1 c_2) \quad (2)$$

**Definition 4.** [46] Let $A_1 = (a_1, b_1, c_1)$ and $A_2 = (a_2, b_2, c_2)$ be two SVN numbers, then multiplication between $A_1$ and $A_2$ is defined as follows:

$$A_1 \otimes A_2 = (a_1 a_2, b_1 + b_2 - b_1 b_2, c_1 + c_2 - c_1 c_2). \quad (3)$$

**Definition 5.** [46] Let $A = (a, b, c)$ be a SVN number and $\lambda \in \mathbb{R}$ an arbitrary positive real number, then

$$\lambda A = (1 - (1-a)^\lambda, b^\lambda, c^\lambda), \lambda > 0. \quad (4)$$

Based on the study given in ([46]), we define the weighted aggregation operators related to SVNSs as follows:

**Definition 6.** Let $\{A_1, A_2, \ldots, A_n\}$ be the set of $n$ SVN numbers, where $A_j = (a_j, b_j, c_j)$ $(j = 1, 2, \ldots, n)$. The single valued neutrosophic weighted average operator on them is defined by

$$\sum_{j=1}^n \lambda_j A_j = \left(1 - \prod_{j=1}^n (1 - a_j)^{\lambda_j}, \left(\prod_{j=1}^n (b_j)^{\lambda_j}, \prod_{j=1}^n (c_j)^{\lambda_j}\right)\right) \quad (5)$$

where $\lambda_j$ is the weight of $A_j$ $(j = 1, 2, \ldots, n)$, $\lambda_j \in [0,1]$ and $\sum_{j=1}^n \lambda_j = 1$.

**Definition 7.** [45] Let $A^* = (A_1^*, A_2^*, \ldots, A_n^*)$ be a vector of $n$ SVN numbers such that $A_j^* = (a_j^*, b_j^*, c_j^*)$ $(j = 1, 2, \ldots, n)$ and $B_i = (B_{i1}, B_{i2}, \ldots, B_{im})$ $(i = 1, 2, \ldots, m)$ be $m$ vectors of $n$ SVN numbers such that $B_{ij} = (a_{ij}, b_{ij}, c_{ij})$ $(i = 1, 2, \ldots, m)$, $(j = 1, 2, \ldots, n)$. Then the separation measure between $B_i$'s and $A^*$ based on Euclidian distance is defined as follows:

$$s_i = \left(\frac{1}{3}\sum_{j=1}^n \left\{\left(|a_{ij} - a_j^*|\right)^2 + \left(|b_{ij} - b_j^*|\right)^2 + \left(|c_{ij} - c_j^*|\right)^2\right\}\right)^{\frac{1}{2}}$$

$$(i = 1, 2, \ldots, m). \quad (6)$$

Next, we proposed a score function for ranking SVN numbers as follows:

**Definition 8.** Let $A = (a, b, c)$ be a single valued neutrosophic number, a score function $S$ of a single valued neutrosophic value, based on the truth-membership degree, indeterminacy-membership degree and falsity membership degree is defined by

$$S(A) = \frac{1 + a - 2b - c}{2} \quad (7)$$

where $S(A) \in [-1, 1]$.

The score function $S$ is reduced the score function proposed by Li ([47]) if $b = 0$ and $a + c \leq 1$.

**Example 9.** Let $A_1 = (0.5, 0.2, 0.6)$ and $A_2 = (0.6, 0.3, 0.2)$ be two single valued neutrosophic numbers for two alternatives. Then, by applying *Definition 8*, we can obtain

$$S(A_1) = \frac{1 + 0.5 - 2 \times 0.2 - 0.6}{2} = 0.25$$

$$S(A_2) = \frac{1 + 0.6 - 2 \times 0.3 - 0.2}{2} = 0.4.$$

In this case, we can say that alternative $A_2$ is better than $A_1$.

## 2.3 TOPSIS Method and Linguistic Variables

In the section, we briefly summarize the TOPSIS method and its applications. Then we discuss the using TOPSIS method in solving MCDM problems. We give the relationships between linguistic variables and single valued neutrosophic numbers.

The TOPSIS (Technique for Order Preference by Similarity to Ideal Solution) method was initiated by Hwang and Yoon ([29]). It is very suitable practical method which is one of the methods of the multi-criteria decision making. In practice, the TOPSIS method is a process of determining the alternative which is closest to the ideal solution, i.e. ranking the alternatives with respect to their distances from the ideal and the negative ideal solution and has applied to many areas relying on computer support to overcome evaluation problems under a finite number of alternatives. In this method, the grades of options are determined according to ideal solution similarity. If the similarity rate of an option is more close to an ideal solution which is the best from any aspect that does not exist practically, it has a higher grade and also is the optimal choice.

A linguistic variable is a variable whose values are characterized with words or sentences instead of numbers in a natural or artificial language. The value of a linguistic variable is expressed as an element of its term set. The concept of a linguistic variable is very useful for solving decision making problems with complex content. For example, we can express the performance ratings of alternatives on qualitative attributes by linguistic variables such as very important, important, medium, unimportant, very unimportant, etc. Such linguistic values can be represented using single valued neutrosophic numbers. For example, 'important' and 'very important' can be expressed by single valued neutrosophic numbers $(0.2, 0.3, 0.5)$ and $(0.6, 0.9, 1.0)$, respectively.

Fundamentally, linguistic terms are individual variations for a linguistic variable. That is, linguistic terms do not meet precise meaning and it may be interpreted differently by different people. The cover of a determined term are pretty subjective and it may vary as the case. Therefore, linguistic terms cannot be indicated by classic set theory and also each linguistic term is associated with a single valued neutrosophic set. The following example illustrates that situation.

**Example 10.** Let $X = \{x_1, x_2, x_3, x_4, x_5\}$ be five alternatives in the universe of cars. Suppose that ''quality of the cars'' is a linguistic variable and $T$(price) = {extremely high, very high, medium, very low} is set of linguistic terms for this variable. Since each linguistic term is characterized with its own single valued neutrosophic set, two of them might be defined as follows:

$$T_{\text{very high}} = \begin{Bmatrix} (x_1, 0.5, 0.7, 0.4), (x_2, 0.1, 0.3, 0.4), \\ (x_4, 0.5, 0.4, 0.1), (x_5, 0.3, 0.3, 0.5) \end{Bmatrix}$$

$$T_{\text{medium}} = \begin{Bmatrix} (x_1, 0.2, 0.4, 0.6), (x_3, 0.3, 0.1, 0.5), \\ (x_4, 0.2, 0.4, 0.7), (x_5, 0.4, 0.1, 0.6) \end{Bmatrix}$$

Supplier selection has a very important place in multi criteria decision making. In our supplier selection approach, we firstly collect the individual evaluations of multiple decision makers and then we decide for a final select. In the method, there are $k$-decision makers, $m$-alternatives and $n$-criteria. $k$-decision makers evaluate the importance of the $m$ alternatives under $n$ criteria and rank the performance of the $n$ criteria with respect to linguistic statements converted into single valued neutrosophic numbers. Here, the decision makers utilize often a set of weights such that $W$ = {very important, important, medium, unimportant, very unimportant} and the importance weights based on single valued neutrosophic values of the linguistic terms is given as Table 1.

**Table 1.** Importance weight as linguistic variables

| Linguistic terms | SVNSs |
| --- | --- |
| Very important (VI) | (0.90, 0.10, 0.10) |
| Important (I) | (0.75, 0.25, 0.20) |
| Medium (M) | (0.50, 0.50, 0.50) |
| Unimportant (UI) | (0.35, 0.75, 0.80) |
| Very unimportant (VUI) | (0.10, 0.90, 0.90) |

Moreover, in Table 2, we give the set of linguistic terms used to rate the importance of alternatives according to decision makers.

**Table 2.** Linguistic terms to rate the importance of alternatives

| Linguistic terms | SVNSs |
| --- | --- |
| Extremely good (EG) / extremely high (EH) | (1.00, 0.00, 0.00) |
| Very very good (VVG) / very very high (VVH) | (0.90, 0.10, 0.10) |
| Very good (VG) / very high (VH) | (0.80, 0.15, 0.20) |
| Good (G) / high (H) | (0.70, 0.25, 0.30) |
| Medium good (MG) / medium high (MH) | (0.60, 0.35, 0.40) |
| Medium (M) / fair (F) | (0.50, 0.50, 0.50) |
| Medium bad (MB) / medium low (ML) | (0.40, 0.65, 0.60) |
| Bad (B) / low (L) | (0.30, 0.75, 0.70) |
| Very bad (VB) / very low (VL) | (0.20, 0.85, 0.80) |

| Very very bad (VVB) / very very low (VVL) | (0.10,0.90,0.90) |
| Extremely bad (EB) / extremely low (EL) | (0.00,1.00,1.00) |

## 3. Single Valued Neutrosophic TOPSIS

Here, we extend the TOPSIS method in single valued neutrosophic sets. Suppose that $A = \{\rho_1, \rho_2, \ldots, \rho_m\}$ is a set of alternatives and $G = \{\beta_1, \beta_2, \ldots, \beta_n\}$ is a set of criteria.

We construct the procedure of single valued neutrosophic TOPSIS process, which is as follows:

*Step 1: Determine the weight of decision makers.*

In the step, we identify the importance of decision-makers using the linguistic set given in Table 1. Assume that our decision group process has $k$ decision makers and $A_t = (a_t, b_t, c_t)$ is a SVN number expressing $t$th decision maker. Then we obtain the weight of $t$th decision maker as follows:

$$\delta_t = \frac{a_t + b_t \left(\frac{a_t}{a_t + c_t}\right)}{\sum_{t=1}^{k} a_t + b_t \left(\frac{a_t}{a_t + c_t}\right)} \qquad (8)$$

$\delta_t \geq 0$ and $\sum_{t=1}^{k} \delta_t = 1$.

Here, the weight of each decision maker is calculated taking into account the truth-membership value, the indeterminacy-membership value and the falsity-membership value from them.

*Step 2: Construction of aggregated single valued neutrosophic decision matrix with respect to decision makers.*

To construct one group decision by aggregating all the individual decisions, we need to obtain aggregated single valued neutrosophic decision matrix $D$. Here, it is defined by $D = \sum_{t=1}^{k} \delta_t D^t$, where $D = d_{ij} = (u_{ij}, r_{ij}, v_{ij})$ and

$$d_{ij} = \langle \left(1 - \prod_{t=1}^{k}(1 - u_{ij}^{(t)})^{\delta_t}, \prod_{t=1}^{k}(r_{ij}^{(t)})^{\delta_t}, \prod_{t=1}^{k}(v_{ij}^{(t)})^{\delta_t}\right) \rangle \qquad (9)$$

Then aggregated single valued neutrosophic decision matrix $D$ of decision makers can be expressed as

$$D = \begin{pmatrix} \rho_{11} & \rho_{12} & \cdots & \rho_{1n} \\ \rho_{21} & \rho_{22} & \cdots & \rho_{2n} \\ \vdots & \vdots & \ddots & \vdots \\ \rho_{m1} & \rho_{m2} & \cdots & \rho_{mn} \end{pmatrix}$$

where $\rho_{ij}$ $(i = 1,2, \ldots, m; j = 1,2, \ldots, n)$ denotes an SVN value.

*Step 3: Determine the weights of criterion.*

Each criteria according to decision makers in decision making process may have different importance. By aggregating the criteria values and the weight values of decision makers for the importance of each criteria, we can obtain the weights of the criteria. Assume that weights of criteria is denoted by $W = (w_1, w_2, \ldots, w_n)$ where $w_j$ indicates the relative importance of criterion $\beta_j$. Let $w_j^{(t)} = \left(a_j^{(t)}, b_j^{(t)}, c_j^{(t)}\right)$ be a SVN number expressing the criteria $\beta_j$ $(j = 1,2, \ldots, n)$ by the $t$th decision maker. Then the weight vector of criteria are obtained by formula (5) as follows:

$$w_j = \delta_1 w_j^{(1)}, \delta_2 w_j^{(2)}, \cdots, \delta_k w_j^{(k)}$$
$$= \langle \left(1 - \prod_{t=1}^{k}(1 - a_j^{(t)})^{\delta_t}, \prod_{t=1}^{k}(b_j^{(t)})^{\delta_t}, \prod_{t=1}^{k}(c_j^{(t)})^{\delta_t}\right) \rangle \qquad (10)$$

*Step 4: Construction of aggregated weighted single valued neutrosophic decision matrix with respect to criteria.*

By using the weight of criteria (W) and the aggregated weighted single valued decision matrix (D), we obtain the aggregated weighted single valued neutrosophic decision matrix. Let us assume that $D^* = (d_{ij}^*)$. Then it is defined by

$$D^* = D \otimes W, \qquad (11)$$

where

$d_{ij}^* = w_j \otimes d_{ij} = (a_{ij}, b_{ij}, c_{ij})$. Thus, the aggregated single valued neutrosophic matrix of criteria can be expressed as

$$D^* = \begin{pmatrix} \rho w_{11} & \rho w_{12} & \cdots & \rho w_{1n} \\ \rho w_{21} & \rho_{22} & \cdots & \rho w_{2n} \\ \vdots & \vdots & \ddots & \vdots \\ \rho w_{m1} & \rho w_{m2} & \cdots & \rho w_{mn} \end{pmatrix}$$

*Step 5: Calculation single valued positive-ideal solution (SVN-PIS) and single valued negative-ideal solution (SVN-NIS).*

In TOPSIS method, the evaluation criteria can be categorized into two categories, benefit and cost. Let $G_1$ be a collection of benefit criteria and $G_2$ be a collection of cost criteria. According to single valued neutrosophic set theory and the principle of classical TOPSIS method, SVN-PIS and SVN-NIS can be defined as follows, respectively;

$$\rho^+ = \left(a_{\rho^+ w}(\beta_j), b_{\rho^+ w}(\beta_j), c_{\rho^+ w}(\beta_j)\right) \quad (12)$$

$$\rho^- = \left(a_{\rho^- w}(\beta_j), b_{\rho^- w}(\beta_j), c_{\rho^- w}(\beta_j)\right) \quad (13)$$

where

$$a_{\rho^+ w}(\beta_j) = \begin{pmatrix} \max_i a_{\rho_i w}(\beta_j), & \text{if } j \in G_1 \\ \min_i a_{\rho_i w}(\beta_j), & \text{if } j \in G_2 \end{pmatrix}$$

$$b_{\rho^+ w}(\beta_j) = \begin{pmatrix} \min_i b_{\rho_i w}(\beta_j), & \text{if } j \in G_1 \\ \max_i b_{\rho_i w}(\beta_j), & \text{if } j \in G_2 \end{pmatrix}$$

$$c_{\rho^+ w}(\beta_j) = \begin{pmatrix} \min_i c_{\rho_i w}(\beta_j), & \text{if } j \in G_1 \\ \max_i c_{\rho_i w}(\beta_j), & \text{if } j \in G_2 \end{pmatrix}$$

and

$$a_{\rho^- w}(\beta_j) = \begin{pmatrix} \min_i a_{\rho_i w}(\beta_j), & \text{if } j \in G_1 \\ \max_i a_{\rho_i w}(\beta_j), & \text{if } j \in G_2 \end{pmatrix}$$

$$b_{\rho^- w}(\beta_j) = \begin{pmatrix} \max_i b_{\rho_i w}(\beta_j), & \text{if } j \in G_1 \\ \min_i b_{\rho_i w}(\beta_j), & \text{if } j \in G_2 \end{pmatrix}$$

$$c_{\rho^- w}(\beta_j) = \begin{pmatrix} \max_i c_{\rho_i w}(\beta_j), & \text{if } j \in G_1 \\ \min_i c_{\rho_i w}(\beta_j), & \text{if } j \in G_2 \end{pmatrix}$$

*Step 6: Calculate of distance measures from SVN-PIS and SVNNIS.*

To measure distance of each alternative $\rho_i$ from SVN-PIS and SVN-NIS, we use the distance measure given by Eq. (6).

$$s_i^+ = \left(\frac{1}{3}\sum_{j=1}^n \left\{(|a_{ij} - a_j^+|)^2 + (|b_{ij} - b_j^+|)^2 + (|c_{ij} - c_j^+|)^2\right\}\right)^{\frac{1}{2}}$$

$$(i = 1, 2, \ldots, m), \quad (14)$$

and

$$s_i^- = \left(\frac{1}{3}\sum_{j=1}^n \left\{(|a_{ij} - a_j^-|)^2 + (|b_{ij} - b_j^-|)^2 + (|c_{ij} - c_j^-|)^2\right\}\right)^{\frac{1}{2}}$$

$$(i = 1, 2, \ldots, m). \quad (15)$$

*Step 7: Calculate the closeness coefficient (CC)*

Finally, we compute relative closeness coefficient of each alternative with respect to single valued neutrosophic ideal solutions by using

$$\tilde{\rho}_j = \frac{s^-}{s^+ + s^-}, \text{ where } 0 \leq \tilde{\rho}_j \leq 1. \quad (16)$$

*Step 8: Determine the rank of alternatives.*

According to descending order of relative closeness coefficient we can rank all alternatives.

## 4. Numerical example

Assume that for supplier selection in a production industry, four decision makers (DM) has been appointed to evaluate 5 supplier alternatives ($\rho_i$; 1, 2, ..., 5) with respect to five performance criteria such that delivery, quality, flexibility, service and price. The decision-makers utilize a linguistic set of weights to determine the performance of each criterion. The information of weights provided to the five criteria by the four decision makers are presented in Table 3.

**Table 3.** The importance weights of the decision criteria

| Criteria | DM(1) | DM(2) | DM(3) | DM(4) |
|---|---|---|---|---|
| Delivery | VI | VI | VI | I |
| Quality | I | M | M | I |
| Flexibility | VI | VI | I | VI |
| Service | I | I | M | UI |
| Price | M | M | VI | VI |

We assume that the decision makers use the linguistic variables and ratings to state the suitability of the supplier alternatives under each of the subjective criteria. The results are shown in Table (4-8.

**Table 4.** The ratings of the alternatives for delivery criterion

| | Delivery | | | |
|---|---|---|---|---|
| Supp. | DM(1) | DM(2) | DM(3) | DM(4) |
| $\rho_1$ | VG | MG | VG | G |
| $\rho_2$ | G | VG | MG | MG |
| $\rho_3$ | M | G | MG | M |
| $\rho_4$ | G | MG | G | MG |
| $\rho_5$ | MG | G | VG | VG |

**Table 5.** The ratings of the alternatives for quality criterion

| | Quality | | | |
|---|---|---|---|---|
| Supp. | DM(1) | DM(2) | DM(3) | DM(4) |
| $\rho_1$ | G | G | MG | G |
| $\rho_2$ | VG | MG | M | MG |
| $\rho_3$ | M | VG | G | G |
| $\rho_4$ | MG | M | VG | M |
| $\rho_5$ | G | G | MG | VG |

**Table 6.** The ratings of the alternatives for flexibility criterion

| | Flexibility | | | |
|---|---|---|---|---|
| Supp. | DM(1) | DM(2) | DM(3) | DM(4) |
| $\rho_1$ | MG | MG | M | M |
| $\rho_2$ | VG | G | VG | VG |
| $\rho_3$ | M | G | MG | MG |
| $\rho_4$ | G | MG | G | MG |
| $\rho_5$ | MG | G | VG | G |

**Table 7.** The ratings of the alternatives for service criterion

| | Service | | | |
|---|---|---|---|---|
| Supp. | DM(1) | DM(2) | DM(3) | DM(4) |
| $\rho_1$ | G | M | MG | M |
| $\rho_2$ | VG | VG | M | G |
| $\rho_3$ | MG | MG | MG | MG |
| $\rho_4$ | M | MB | MG | VG |
| $\rho_5$ | MG | G | VG | G |

**Table 8.** The ratings of the alternatives for price criterion

| | Price | | | |
|---|---|---|---|---|
| Supp. | DM(1) | DM(2) | DM(3) | DM(4) |
| $\rho_1$ | M | MH | VH | M |
| $\rho_2$ | VH | M | H | H |
| $\rho_3$ | H | H | M | MH |
| $\rho_4$ | M | M | MH | H |
| $\rho_5$ | H | VH | VH | VH |

Next, we apply the procedure of single valued neutrosophic TOPSIS process, which is as follows:

**Step 1**. Determine the weights of the decision makers. By using Eq. (8), we obtain the weights of the decision makers (Table 9).

**Table 9.** The importance of decision makers and their weights.

| | DM(1) | DM(2) | DM(3) | DM(4) |
|---|---|---|---|---|
| Ling. T. | VI | I | M | UI |
| Weight | 0.2864 | 0.2741 | 0.2170 | 0.1673 |

Then we denotes the weight vector of the decision makers by $\delta = [\delta_1, \delta_2, \delta_3, \delta_4]$.

**Step 2.** Construction of aggregated single valued neutrosophic decision matrix with respect to decision makers.

The ratings assigned by the decision makers to each alternative were given in Table (4-8), respectively. Then the aggregated SVN decision matrix obtained by aggregating of opinions of decision makers is constructed by Eq. (9). The result is given in Table 10.

**Step 3.** Determine the weights of criterion.

We calculate the weights of each criterion by using Eq. (10). In order to do that, we use the information from Table 3 and present it in Table 12.

**Table 10.** Aggregated SVN decision matrix

| | $\beta_1$ | $\beta_2$ | $\beta_3$ | $\beta_4$ | $\beta_5$ |
|---|---|---|---|---|---|
| $\rho_1$ | (0.717,0.228,0.282) | (0.658,0.290,0.341) | (0.541,0.425,0.458) | (0.572,0.394,0.427) | (0.599,0.362,0.400) |
| $\rho_2$ | (0.679,0.266,0.320) | (0.637,0.314,0.362) | (0.755,0.191,0.244) | (0.714,0.235,0.285) | (0.671,0.281,0.328) |
| $\rho_3$ | (0.548,0.429,0.451) | (0.651,0.302,0.348) | (0.585,0.374,0.414) | (0.579,0.370,0.420) | (0.624,0.348,0.375) |
| $\rho_4$ | (0.636,0.313,0.363) | (0.600,0.361,0.399) | (0.636,0.313,0.363) | (0.553,0.422,0.446) | (0.545,0.428,0.454) |
| $\rho_5$ | (0.702,0.244,0.297) | (0.681,0.266,0.318) | (0.681,0.265,0.318) | (0.681,0.279,0.318) | (0.754,0.192,0.245) |

**Table 11.** Aggregated weighted SVN decision matrix

| | $\beta_1$ | $\beta_2$ | $\beta_3$ | $\beta_4$ | $\beta_5$ |
|---|---|---|---|---|---|
| $\rho_1$ | (0.622,0.330,0.374) | (0.571,0.384,0.425) | (0.469,0.474,0.500) | (0.496,0.474,0.500) | (0.520,0.447,0.476) |
| $\rho_2$ | (0.421,0.544,0.553) | (0.396,0.574,0.580) | (0.469,0.525,0.530) | (0.443,0.525,0.530) | (0.416,0.554,0.558) |
| $\rho_3$ | (0.472,0.508,0.523) | (0.561,0.399,0.434) | (0.504,0.457,0.497) | (0.499,0.457,0.497) | (0.537,0.438,0.458) |
| $\rho_4$ | (0.402,0.571,0.577) | (0.379,0.601,0.601) | (0.402,0.640,0.632) | (0.349,0.640,0.632) | (0.344,0.643,0.637) |
| $\rho_5$ | (0.505,0.455,0.497) | (0.490,0.471,0.509) | (0.490,0.481,0.509) | (0.490,0.481,0.509) | (0.543,0.418,0.456) |

**Table 12.** The weights of criteria.

| Criteria | Weight |
|---|---|
| $\beta_1$ | (0.867,0.132,0.127) |
| $\beta_2$ | (0.620,0.379,0.342) |
| $\beta_3$ | (0.861,0.138,0.131) |
| $\beta_4$ | (0.632,0.367,0.336) |
| $\beta_5$ | (0.720,0.279,0.279) |

**Table 13.** SVN- PIS and SVN-NIS values.

| | SVN PIS | SVN NIS |
|---|---|---|
| $\beta_1$ | (0.622,0.330,0.374) | (0.402,0.571,0.577) |
| $\beta_2$ | (0.571,0.384,0.425) | (0.379,0.601,0.601) |
| $\beta_3$ | (0.504,0.457,0.497) | (0.402,0.640,0.632) |
| $\beta_4$ | (0.499,0.457,0.497) | (0.349,0.640,0.632) |
| $\beta_5$ | (0.344,0.643,0.637) | (0.543,0.418,0.456) |

**Step 4:** Construction of aggregated weighted single valued neutrosophic decision matrix with respect to criteria.

To construct the aggregated weighted SVN decision matrix, we use the Eq. (11) and give it in Table 11.

**Step 5.** Calculation SVN positive-ideal solution and SVN negative-ideal solution.

By using Eq. (12) and Eq. (13), SVN positive-ideal solution and SVN negative-ideal solution were calculated as Table 13.

**Step 6.** Calculate the separation measures.

Separation measure of each alternative from the positive-ideal solution and negative ideal solution are calculated using Eq. (14) and Eq. (15) and are given by Table 14.

**Table 14.** Separation measures and the relative closeness coefficient of each alternative.

| Alter. | $d^-$ | $d^+$ | $CC$ | Ranking |
|---|---|---|---|---|
| $\rho_1$ | 0.016 | 0.063 | 0.797 | 1 |
| $\rho_2$ | 0.040 | 0.018 | 0.307 | 4 |
| $\rho_3$ | 0.031 | 0.041 | 0.570 | 2 |
| $\rho_4$ | 0.066 | 0.020 | 0.235 | 5 |
| $\rho_5$ | 0.031 | 0.029 | 0.483 | 3 |

**Step 7:** Calculate the closeness coefficient (CC)

We determine the closeness coefficient of all alternative by Eq. (16). The last column of Table 14 presents the result.

**Step 8.** Rank the alternatives.

According to descending order of relative closeness coefficients values, four alternatives are ranked as $\rho_1 > \rho_3 > \rho_5 > \rho_2 > \rho_4$ as in Table 14. Then, the alternative $\rho_1$ is also the most desirable alternative.

From the example, we can see that the proposed neutrosophic decision-making method is more suitable for real scientific and engineering applications because it can handle not only incomplete information but also the inconsistent information and indeterminate information existing in real world. The technique proposed in this paper extends existing decision making methods and provides a new viewpoint for multi criteria group decision making.

The TOPSIS method is a very important technical for the process of multi criteria decision making. There are many TOPSIS methods for solving multi criteria decision making problems with the fuzzy information and its extension, the intuitionistic fuzzy information and the interval valued intuitionistic fuzzy information. Since the single valued neutrosophic sets generalize the concepts of fuzzy sets and intuitionistic fuzzy sets, the existing TOPSIS methods is not suitable for handling the single valued neutrosophic information including unknown weights of decision makers and the criteria values for alternatives. Therefore, we need to extend the method to neutrosophic environment. The developed decision making method can utilize the proposed score function of single valued neutrosophic numbers to rank alternatives in the process of multi criteria decision making. The performance ratings of decision makers and criteria on alternatives are characterized by linguistic variables. The weights of decision makers are calculated via a developed equation while the weights of the criteria are obtained by aggregating the criteria values provided by decision makers and the weight values of decision makers for the importance of each criteria. The method proposed in this paper are general and more flexible than existing decision making methods.

## 5. Conclusions

In this paper, we extended TOPSIS method that is one of the familiar methods in multi-attribute decision-making problem in single valued neutrosophic sets and proposed a multi-criteria group decision making based on single valued neutrosophic TOPSIS for evaluation of supplier. Since to solve a decision making problem expressed by crisp data is more difficult under uncertain environment, single valued neutrosophic sets are more useful to overcome such situations. In the evaluation process, weights of decision makers, the aggregation of the criteria and the impact of alternatives on criteria with respect to decision makers is very important to appropriately perform evaluation process. In order to do that, the ratings of each alternative according to each criterion and the weights of each criterion were provided as linguistic terms expressed by single valued neutrosophic numbers. Also SVNWA operator is utilized to aggregate all individual decision makers' opinions for determining the importance of criteria and the alternatives. Firstly, single valued neutrosophic positive-ideal solution and single valued neutrosophic negative-ideal solution were obtained using the Euclidean distance. Then the relative closeness coefficients of alternatives were calculated and finally ranking the alternatives was done.

TOPSIS method based on single valued neutrosophic set is more useful for solving multi-criteria decision-making problems because of considering order of importance of decision makers. So, the single valued neutrosophic TOPSIS can be preferable for dealing with incomplete, indetermine and inconsistent information in MCDM problems such as selecting project and personnel, selecting a flexible manufacturing system and many further areas of marketing research problems and management decision problems.

The relationships among the alternatives and their positive-ideal solution and negative ideal solution is presented in Fig. (1-5).

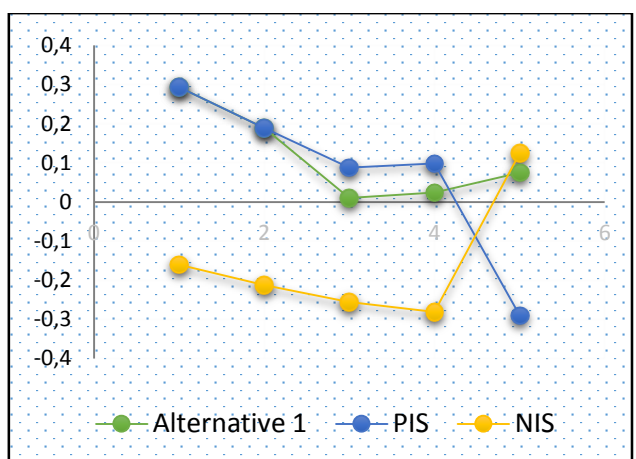

**Figure 1:** The relationships among $A_1$ and its PIS and NIS

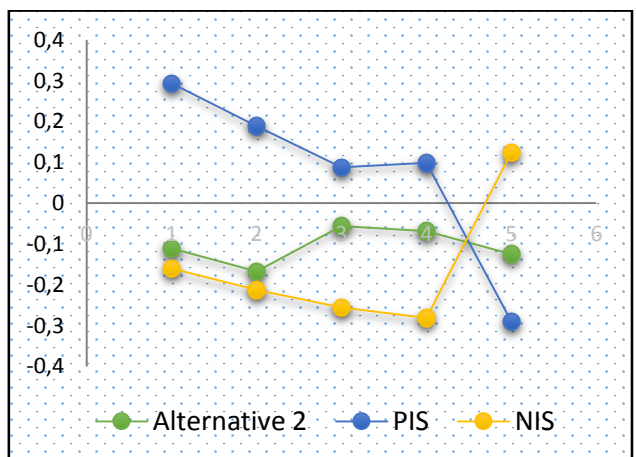

**Figure 2:** The relationships among $A_2$ and its PIS and NIS

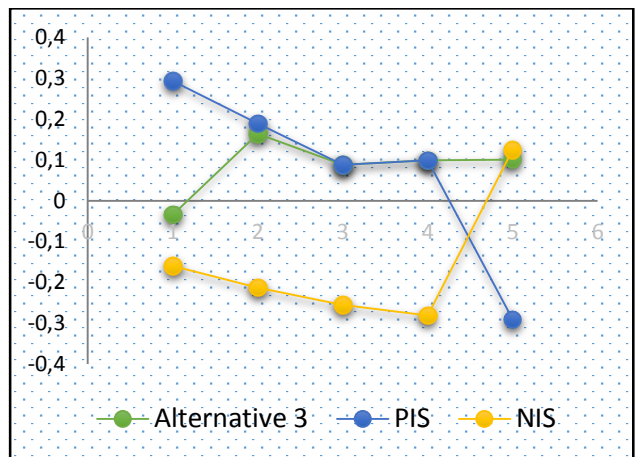

**Figure 1:** The relationships among $A_3$ and its PIS and NIS

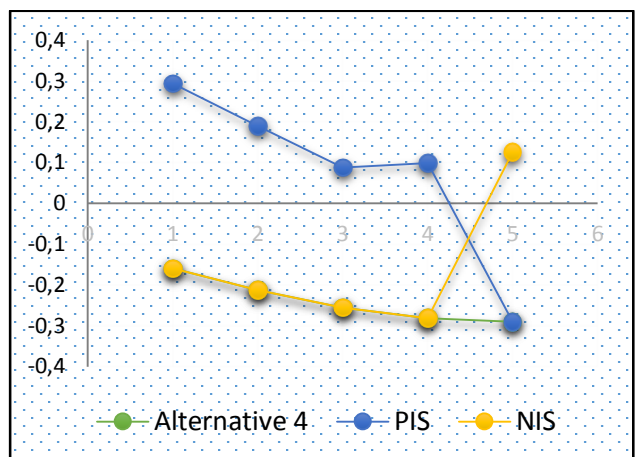

**Figure 4:** The relationships among $A_4$ and its PIS and NIS

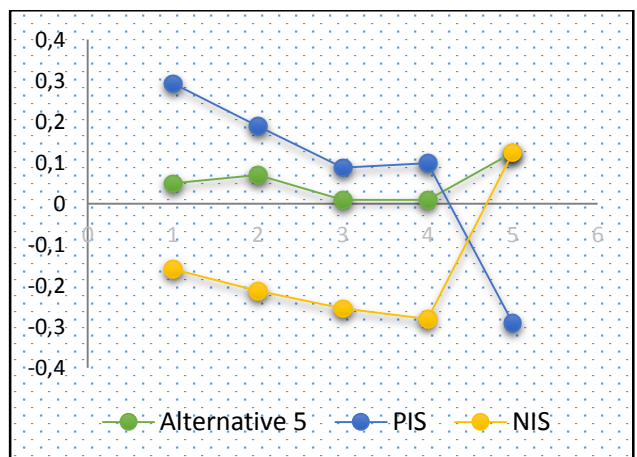

**Figure 5:** The relationships among $A_5$ and its PIS and NIS

## References

[1] Smarandache, F. (1999). A unifying field in logics. Neutrosophy: Neutrosophic probability, set and logic, American Research Press, Rehoboth.

[2] Smarandache, F. (2005). A generalization of the intuitionistic fuzzy set. International journal of Pure and Applied Mathematics, 24, 287-297.

[3] L. A. Zadeh, "Fuzzy sets," Information and Control, vol. 8, no.3, pp. 338–353, 1965.

[4] K. T. Atanassov, "Intuitionistic fuzzy sets," Fuzzy Sets and Systems, vol. 20, no. 1, pp. 87–96, 1986.

[5] K. T. Atanassov and G. Gargov, "Interval valued intuitionistic fuzzy sets," Fuzzy Sets and Systems, vol. 31, no. 3, pp. 343–349, 1989

[6] Wang, H., Smarandache, F., Zhang, Y. Q., and Sunderraman, R., (2010. Single valued neutrosophic sets, Multispace and Multistructure (4) 410-413

[7] Wang, H., Smarandache, F., Zhang, Y. Q. and Sunderraman. (2005). Interval neutrosophic sets and logic: Theory and applications in computing, Hexis, Phoenix, AZ.

[8] S. H. Ha and R. Krishnan, "A hybrid approach to supplier selection for the maintenance of a competitive supply chain," Expert Systems with Applications, vol. 34, no. 2, pp. 1303–1311, 2008.

[9] L. M. Ellram, "The supplier selection decision in strategic partnerships," Journal of Purchasing Material Management, vol. 26, no. 4, pp. 8–14, 1990.

[10] C. Araz and I. Ozkarahan, "Supplier evaluation and management system for strategic sourcing based on a new multicriteria sorting procedure," International Journal of Production Economics, vol. 106, no. 2, pp. 585–606, 2007.

[11] L. de Boer, L. van der Wegen, and J. Telgen, "Outranking methods in support of supplier selection," European Journal of Purchasing and Supply Management, vol. 4, no. 2-3, pp. 109–118, 1998.

[12] H. Min, "International supplier selection: a multi-attribute utility approach," International Journal of Physical Distribution and Logistics Management, vol. 24, no. 5, pp. 24–33, 1994.

[13] E. Timmerman, "An approach to vendor performance evaluation," Journal of Purchasing and Supply Management, vol. 1, pp. 27–32, 1986.

[14] R. E. Gregory, "Source selection: a matrix approach," Journal of Purchasing and Materials Management, vol. 22, no. 2, pp. 24–29, 1986.

[15] W. R. Soukup, "Supplier selection strategies," Journal of Purchasing and Materials Management, vol. 23, no. 3, pp. 7–12, 1987.

[16] K. Thompson, "Vendor prole analysis," Journal of Purchasing and Materials Management, vol. 26, no. 1, pp. 11–18, 1990.

[17] G. Barbarosoglu and T. Yazgac, "An application of the analytic hierarchy process to the supplier selection problem," Production and Inventory Management Journal, vol. 38, no. 1, pp. 14–21, 1997.

[18] R. Narasimhan, "An analytic approach to supplier selection,"Journal of Purchasing and Supply Management, vol. 1, pp. 27–32, 1983.

[19] R. L. Nydick and R. P. Hill, "Using the Analytic Hierarchy Process to structure the supplier selection procedure," International Journal of Purchasing and Materials Management, vol. 28, no. 2, pp. 31–36, 1992.

[20] J. Sarkis and S. Talluri, "A model for strategic supplier selection," in Proceedings of the 9th International IPSERA Conference, M. Leenders, Ed., pp. 652–661, Richard Ivey Business School, London ,UK, 2000.

[21] S. S. Chaudhry, F. G. Forst, and J. L. Zydiak, "Vendor selection with price breaks," European Journal of Operational Research, vol. 70, no. 1, pp. 52–66, 1993.

[22] E. C. Rosenthal, J. L. Zydiak, and S. S. Chaudhry, "Vendor selection with bundling," Decision Sciences, vol. 26, no. 1, pp. 35–48, 1995.

[23] F. P. Bua and W. M. Jackson, "A goal programming model for purchase planning," Journal of Purchasing and Materials Management, vol. 19, no. 3, pp. 27–34, 1983.

[24] S. H. Ghodsypour and C. O. O'Brien, "A decision support system for supplier selection using an integrated analytic hierarchy process and linear programming," International Journal of Production Economics, vol. 56-57, no. 13, pp. 199–212, 1998.

[25] C. A.Weber and L. M. Ellram, "Supplier selection using multi objective programming: a decision support system approach, "International Journal of Physical Distribution and Logistics Management, vol. 23, no. 2, pp. 3–14, 1992.

[26] C. C. Li, Y. P. Fun, and J. S. Hung, "A new measure for supplier performance evaluation," IIE Transactions on Operations Engineering, vol. 29, no. 9, pp. 753–758, 1997.

[27] G. D. Holt, "Which contractor selection methodology?" International Journal of Project Management, vol. 16, no. 3, pp. 153–164, 1998.

[28] C. T. Chen, C. T. Lin, and S. F. Huang, "A fuzzy approach for supplier evaluation and selection in supply chain management," International Journal of Production Economics, vol. 102, no. 2, pp. 289–301, 2006.

[29] C. L. Hwang and K. Yoon, Multiple Attribute Decision Making Methods and Applications, Springer, Heidelberg, Germany, 1981.

[30] A. N. Haq and G. Kannan, "Fuzzy analytical hierarchy process for evaluating and selecting a vendor in a supply chainmodel, "International Journal of Advanced Manufacturing Technology, vol. 29, no. 7-8, pp. 826–835, 2006.

[31] M. Y. Bayrak, N. Celebi, and H. Takin, "A fuzzy approach method for supplier selection," Production Planning and Control: The Management of Operations, vol. 18, no. 1, pp. 54–63, 2007.

[32] F. T. S. Chan, N. Kumar, M. K. Tiwari, H. C. W. Lau, and K. L. Choy, "Global supplier selection: a fuzzy-AHP approach," International Journal of Production Research, vol. 46, no. 14, pp. 3825–3857, 2008.

[33] S. Önüt, S. S. Kara, and E. Işık, "Long term supplier selection using a combined fuzzy MCDM approach: a case study for a telecommunication company," Expert Systems with Applications, vol. 36, no. 2, pp. 3887–3895, 2009.

[34] S. J. Chen and C. L. Hwang, Fuzzy Multiple Attribute Decision Making: Methods and Applications, Springer, Berlin, Germany, 1992.

[35] S. H. Tsaur, T. Y. Chang, and C. H. Yen, "The evaluation of airline service quality by fuzzy MCDM," Tourism Management, vol. 23, pp. 107–115, 2002.

[36] T. C. Chu, "Selecting plant location via a fuzzy TOPSIS approach," International Journal of Advanced Manufacturing Technology, vol. 20, no. 11, pp. 859–864, 2002.

[37] C. T. Chen, "Extensions of the TOPSIS for group decision making under fuzzy environment," Fuzzy Sets and Systems, vol. 114, no. 1, pp. 1–9, 2000.

[38] G. R. Jahanshahloo, F. H. Lotfi, and M. Izadikhah, "An algorithmic method to extend TOPSIS for decision-making problems with interval data," Applied Mathematics and Computation, vol. 175, no. 2, pp. 1375–1384, 2006.

[39] G. R. Jahanshahloo, F. H. Lotfi, and M. Izadikhah, "Extension of the TOPSIS method for decision-making problems with fuzzy data," Applied Mathematics and Computation, vol. 181, no. 2, pp. 1544–1551, 2006.

[40] M. Izadikhah, "Using the Hamming distance to extend TOPSIS in a fuzzy environment," Journal of Computational and Applied Mathematics, vol. 231, no. 1, pp. 200–207, 2009.

[41] F. E. Boran, S. Genc,, M. Kurt, and D. Akay, "A multi-criteria intuitionistic fuzzy group decision making for supplier selection with TOPSIS method," Expert Systems with Applications, vol. 36, no. 8, pp. 11363–11368, 2009.

[42] F. Ye, "An extended TOPSIS method with interval-valued intuitionistic fuzzy numbers for virtual enterprise partner

selection," Expert Systems with Applications, vol. 37, no. 10, 2010.
[43] J. Ye, Multicriteria decision-making method using the correlation coefficient under single-valued neutrosophic environment, International Journal of General Systems 42(4) (2013), 386-394.
[44] J. Ye, Single valued neutrosophic cross-entropy for multi criteria decision making problems, Applied Mathematical Modelling (2013) doi:10.1016/j.apm.2013.07.020.
[45] J. Ye, Single-Valued Neutrosophic Minimum Spanning Tree and Its Clustering Method, Journal of Intelligent Systems, (2014) DOI: 10.1515/jisys-2013-0075.
[46] Z. Hongyu, J. Qiang Wang, and X. Chen, Interval Neutrosophic Sets and its Application in Multi-criteria Decision Making Problems, The Scientific World Journal, to appear (2013).
[47] Li, D.-F. (2005). Multiattribute decision making models and methods using intuitionistic fuzzy sets. Journal of Computer and System Sciences, 70, 73–85.